\ifdefined\pdfoutput
  \pdfoutput=1
\fi

\documentclass[11pt]{article}

\usepackage[margin=1in]{geometry}

\usepackage[T1]{fontenc}
\usepackage[utf8]{inputenc}
\usepackage{lmodern}
\usepackage{textcomp}

\ifdefined\DeclareUnicodeCharacter
  \DeclareUnicodeCharacter{2013}{\textendash}         
  \DeclareUnicodeCharacter{2014}{\textemdash}         
  \DeclareUnicodeCharacter{2011}{-}                   
  \DeclareUnicodeCharacter{2212}{-}                   
  \DeclareUnicodeCharacter{00B0}{\textdegree}         
  \DeclareUnicodeCharacter{00B2}{\textsuperscript{2}} 
  \DeclareUnicodeCharacter{2019}{\textquotesingle}    
  \DeclareUnicodeCharacter{201C}{``}                  
  \DeclareUnicodeCharacter{201D}{''}                  
  \DeclareUnicodeCharacter{2026}{\ldots}              
\fi

\usepackage{amsmath,amssymb,amsfonts}
\usepackage{booktabs}
\usepackage{microtype}
\usepackage{graphicx}
\usepackage{xcolor}
\usepackage{siunitx}
\usepackage{enumitem}
\setlist[enumerate]{leftmargin=*,labelsep=0.5em}
\usepackage{multirow}
\usepackage{longtable}
\usepackage{array}
\usepackage[most]{tcolorbox}
\usepackage{listings}
\usepackage{float}

\lstdefinestyle{mystyle}{
    basicstyle=\ttfamily\footnotesize,
    backgroundcolor=\color{gray!15},
    frame=none,
    breaklines=true,
    showstringspaces=false,
    tabsize=4,
    literate=
      {–}{{-}}1       
      {—}{{--}}1      
      {‐}{{-}}1       
      {‑}{{-}}1       
      {−}{{-}}1       
      {’}{{'}}1       
      {“}{{``}}1      
      {”}{{''}}1      
      {…}{{\ldots}}1 
      {°}{{\textdegree}}1
      {²}{{\textsuperscript{2}}}1
}

\usepackage{natbib}
\setcitestyle{authoryear,round,semicolon}

\usepackage{hyperref}
\hypersetup{colorlinks=true, linkcolor=black, citecolor=black, urlcolor=black}

\newcommand{\keywords}[1]{\vspace{0.5em}\noindent\textbf{Keywords—} #1}
\newcommand{\Description}[1]{}

\title{Do Vision--Language Models See Urban Scenes as People Do? An Urban Perception Benchmark}

\author{Rashid Mushkani\\
Université de Montréal\\
Mila -- Quebec AI Institute}

\date{} 

\begin{document}
\maketitle

\begin{abstract}
Understanding how people read city scenes can inform design and planning. We introduce a small benchmark for testing vision–language models (VLMs) on urban perception using 100 Montreal street images, evenly split between photographs and photorealistic synthetic scenes. Twelve participants from seven community groups supplied 230 annotation forms across 30 dimensions mixing physical attributes and subjective impressions. French responses were normalized to English. We evaluated seven VLMs in a zero-shot setup with a structured prompt and deterministic parser. We use accuracy for single-choice items and Jaccard overlap for multi-label items; human agreement uses Krippendorff's $\alpha$ and pairwise Jaccard. Results suggest stronger model alignment on visible, objective properties than subjective appraisals. The top system (\texttt{Claude-sonnet}) reaches macro 0.31 and mean Jaccard 0.48 on multi-label items. Higher human agreement coincides with better model scores. Synthetic images slightly lower scores. We release the benchmark, prompts, and harness for reproducible, uncertainty-aware evaluation in participatory urban analysis.

\vspace{0.5cm}

\noindent{Dataset is available: https://huggingface.co/datasets/rsdmu/urban-perception-benchmark}

\end{abstract}


\section{Introduction}

Cities are interpreted through perception as well as through built form. Foundational urban design scholarship examined legibility and the imageability of place \citep{lynch1960image}, whereas urban computing has moved toward large-scale, data-driven accounts of city life \citep{zheng2014urban}. Perception is difficult to measure at scale. Pairwise comparisons of street images revealed stable aggregate judgments of safety and appeal across cities \citep{salesses2013collaborative}, and follow-up datasets such as Place Pulse 2.0 enabled learning models that approximate crowd judgments \citep{dubey2016placepulse}. In parallel, computer vision on street-level imagery uncovered relationships between visual cues and indicators such as socioeconomic change, urban form, or mobility \citep{fan2023urban,jean2016poverty,yeh2020africa,tatem2017worldpop}.

Recent vision--language models (VLMs) integrate image understanding with natural language reasoning \citep{achiam2023gpt4,li2023blip2,dai2023instructblip,llava_next,google2023gemini}. They promise general, instruction-following analysis of images without task-specific training. However, there is limited evidence about how well such systems emulate human perception of urban scenes, particularly on subjective qualities. Existing evaluations of spatial knowledge in language models highlight both competence and brittleness \citep{gurnee2023spacetime,roberts2023gpt4geo,mai2023geoai,bhandari2023geospatial,manvi2024llmgeobias}. Bringing these strands together requires human-centered protocols that respect subjectivity, document uncertainty, and make evaluation reproducible.

\paragraph{Research Questions:}
\begin{itemize}
    \item \textbf{RQ1} To what extent do VLMs align with human judgments on \emph{objective} (visually grounded) versus \emph{subjective} (appraisal) dimensions?
    \item \textbf{RQ2} How does inter-annotator reliability relate to model alignment across dimensions?
    \item \textbf{RQ3} How do results differ between photographic and photorealistic synthetic scenes?
\end{itemize}

\paragraph{Contributions.} We contribute: (1) a documented, community-grounded benchmark of 100 Montreal street-level scenes with \emph{30} perception dimensions and deterministic French--English normalization; (2) a transparent, reproducible zero-shot evaluation harness for seven VLMs; and (3) analyses addressing RQ1--RQ3, including human reliability, subjective vs.~objective performance, and real–synthetic differences.

\section{Related Work}

\paragraph{Urban perception at scale.}
Pairwise comparisons facilitated large-scale mapping of perceived urban qualities, starting with the ``collaborative image of the city'' \citep{salesses2013collaborative} and extended by Place Pulse 2.0 \citep{dubey2016placepulse}. Street-view imagery has subsequently supported tasks from estimating socioeconomic status to tracking urban change \citep{fan2023urban}. At regional scales, multispectral imagery and machine learning have been used to reveal socioeconomic gradients \citep{jean2016poverty,yeh2020africa,tatem2017worldpop}. This body of work suggests that images carry strong signals of the built environment and social life while underscoring that perception is not reducible to a single ground truth \citep{mushkani2025streetreviewparticipatoryaibased}.

\paragraph{Vision--language models and spatial knowledge.}
VLMs unify visual perception and language modeling \citep{achiam2023gpt4,li2023blip2,dai2023instructblip,llava_next,google2023gemini}. Tooling for systematic evaluation is emerging \citep{duan2024vlmevalkit}. Separate threads probe whether language models encode spatial and geographic structure \citep{gurnee2023spacetime,roberts2023gpt4geo,mai2023geoai,bhandari2023geospatial,manvi2024llmgeobias}. The present work examines the intersection: how well do VLMs reproduce human appraisals of urban scenes when provided with a structured prompt and held to a reproducible scoring protocol?

\section{Dataset, Participants, and Annotation Protocol}

\paragraph{Image panels and sources.}
We curated 100 street-level scenes in Montreal. Images are organized into ten panels ($p1{\ldots}p10$), each containing ten scenes. Panels $p1$--$p5$ consist of photorealistic \emph{synthetic} images, while $p6$--$p10$ contain \emph{photographs} (50 images per source type). Synthetic scenes were screened for plausibility and filtered to avoid artifacts that would dominate perception. Across panels we sought variety in locations, times of day, presence of vegetation, and crowding.

\begin{figure}[H]
  \centering
  \includegraphics[height=0.35\textheight]{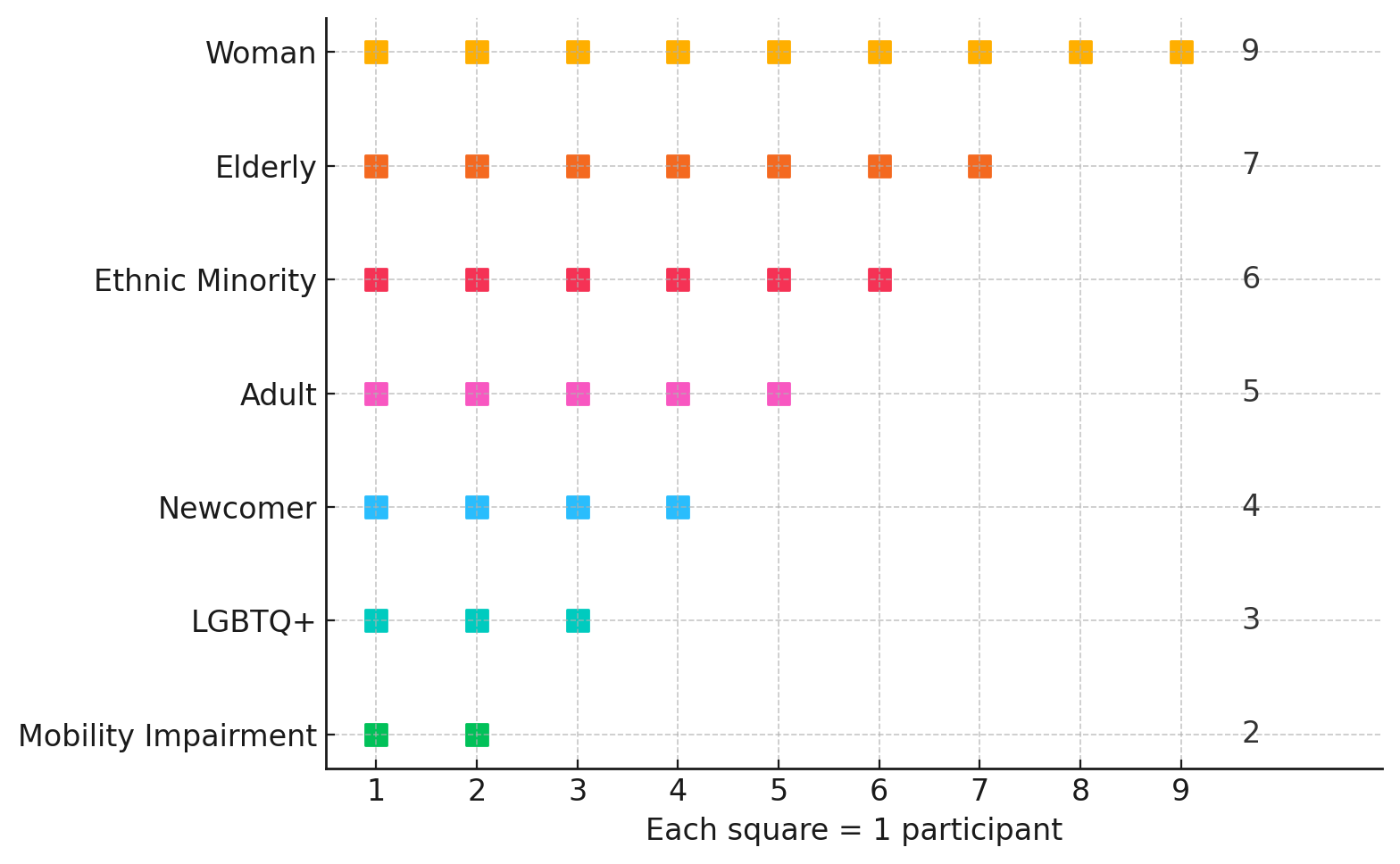}
  \caption{Self-reported participant context (counts). These categories were optional and are reported only in aggregate to characterize the annotator pool. Categories are not mutually exclusive and reflect intersectional identities; participants could select multiple identity markers, so counts exceed the number of participants.}
  \Description{Horizontal bar chart showing counts for Adult, Elderly, Woman, LGBTQ+, Mobility Impairment, Newcomer, and Ethnic Minority.}
  \label{fig:participants}
\end{figure}

\paragraph{Participants and recruitment.}
Twelve Montreal-based participants from seven community organizations annotated the images. Participants were recruited via partner organizations and compensated for their time. Self-identification for context was optional. Figure~\ref{fig:participants} summarizes the diversity of the participant pool.\footnote{Categories were self-reported and optional; no individual-level demographics are released.}

\paragraph{Dimensions.}
The annotation schema comprises \textbf{30} dimensions spanning four families: physical setting (e.g., space typology, spatial configuration, lighting, vegetation, maintenance, signage, barriers), human presence and activity (e.g., human presence, types of activities, economic activities, accessibility features, visibility), built form and aesthetics (e.g., built environment, architectural style, aesthetic elements, cultural elements), and subjective impressions (e.g., overall impression). Each dimension was defined with concise descriptions and canonical label sets, distinguishing single-choice from multi-label items. To avoid identity inference, we use observational descriptors such as \emph{Presence of mobility aids} rather than attributes like ethnicity or gender.

\paragraph{Collection.}
Each image received between one and three independent annotations, resulting in 230 completed forms. A shared subset ensured overlap across participants for reliability analysis. Annotation was conducted in French.

\paragraph{Normalization and consensus.}
We constructed a deterministic French--English mapping for each possible response string, covering synonyms and capitalization. For multi-label dimensions, we aggregated selections per image and formed a \emph{hard consensus set} by retaining options selected by at least half of annotators ($\ge 50\%$). For single-choice dimensions, we used majority vote; exact ties were flagged \emph{unsure} and excluded from accuracy calculations. To avoid inflated agreement, ``Not applicable'' was not rewarded as a match on multi-label Jaccard scores; empty human and model sets were treated as missing rather than perfect overlap.

\section{Models and Zero-Shot Protocol}

\paragraph{Evaluated models.}
We evaluate seven VLMs representative of current practice: \texttt{claude-sonnet}, \texttt{openai-o4-mini}, \texttt{gpt-4.1}, \texttt{gemini-2.5-pro}, \texttt{grok-2-vision}, \texttt{qwen2.5-vl}, and \texttt{llama-4-maverick}. All models were queried through the same script that embeds local images as base-64 data URIs (ensuring pixel access) and requests a single-line CSV response. For determinism we set \texttt{temperature=0}, \texttt{top\_p=1}, and log model version strings and run dates.

\paragraph{Prompting and parsing.}
The prompt enumerates the 30 dimensions in a fixed order with short definitions. It asks the model to choose one label for single-choice items and any number of labels for multi-label items. The script enforces a strict CSV format and retries on parse failures. The parser is rule-based and deterministic: it tokenizes the model CSV, trims whitespace, normalizes spelling, and maps tokens to codebook labels. The pipeline yields complete per-image, per-dimension model responses for all models; non-conforming replies are captured in a \texttt{Comments} column and excluded from scoring.

\section{Evaluation Methodology}

\paragraph{Scoring.}
For single-choice dimensions we compute accuracy against the human majority label, excluding images marked \emph{unsure}. For multi-label dimensions we compute the Jaccard index between the model-selected set and the human hard consensus set, not rewarding ``Not applicable''; images where both sets are empty are treated as missing for that dimension. We then average scores per dimension and report macro-averages across the 30 dimensions. We additionally report set overlap restricted to the multi-label subset to illuminate where selection breadth matters.

\paragraph{Human reliability.}
Inter-annotator reliability is computed per dimension. For single-choice items we use Krippendorff's $\alpha$ with nominal distance; for multi-label items we compute mean pairwise Jaccard across annotator sets, per image, then average per dimension. These measures are not interchangeable but together indicate which dimensions admit more stable human judgments.

\paragraph{Additional analyses.}
We analyze the relationship between human reliability and average model performance across dimensions; we compare performance across subjective and objective subsets of dimensions; and we visualize distributional mismatches for the ``Overall Impression'' dimension. Synthetic-versus-real gaps are computed analogously by stratifying images by source. For multiple variable-level comparisons in exploratory analyses we control the false discovery rate with the Benjamini--Hochberg procedure \citep{benjamini1995fdr}.

\section{Results}

\begin{figure}[H]
  \centering
  \includegraphics[width=\linewidth,height=0.35\textheight,keepaspectratio]{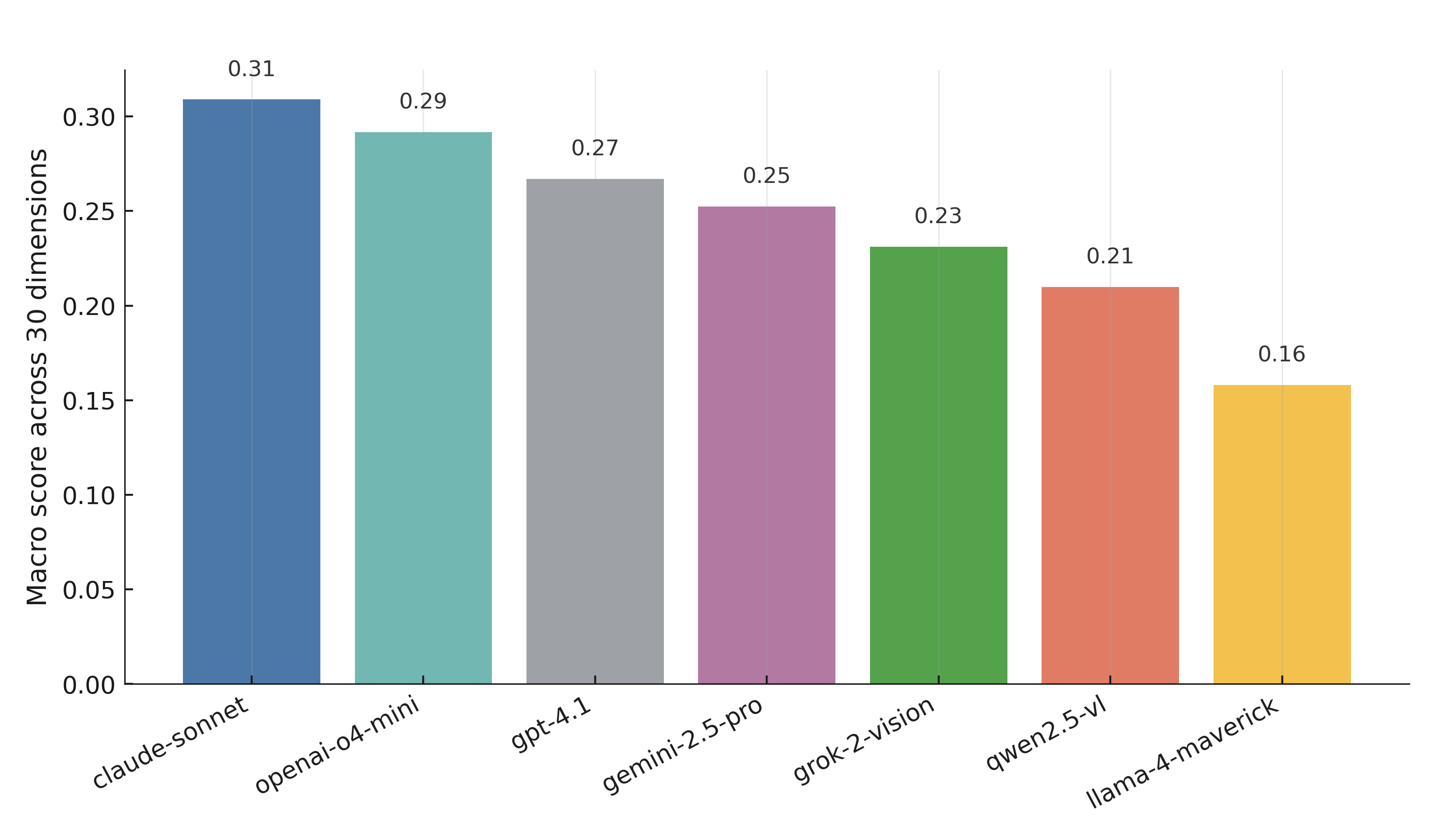}
  \caption{Overall agreement with human consensus by model. Macro-averaged accuracy (single-choice) and Jaccard (multi-label) across 30 dimensions.}
  \Description{Bar chart of overall model alignment. claude-sonnet ≈0.31; openai-o4-mini ≈0.29; gpt-4.1 ≈0.27; gemini-2.5-pro ≈0.25; grok-2-vision ≈0.23; qwen2.5-vl ≈0.21; llama-4-maverick ≈0.16.}
  \label{fig:overall-macro}
\end{figure}

\begin{figure}[H]
  \centering
  \includegraphics[width=\linewidth,height=0.35\textheight,keepaspectratio]{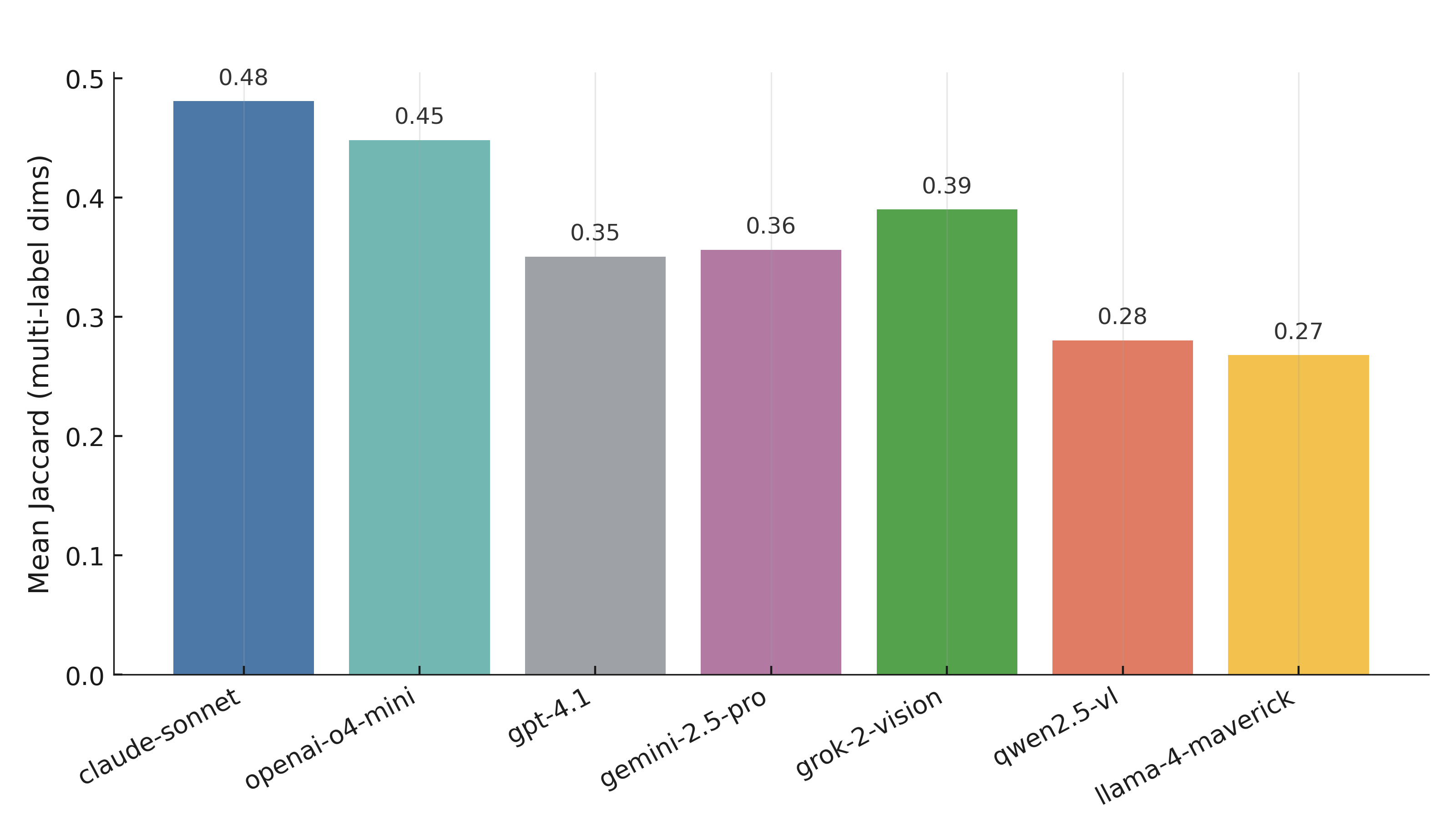}
  \caption{Mean Jaccard set overlap between model selections and the human consensus for the multi-label dimensions.}
  \Description{Bar chart of mean Jaccard on multi-label items. Top two: claude-sonnet ≈0.48 and openai-o4-mini ≈0.45; others lower.}
  \label{fig:overall-jaccard}
\end{figure}

\paragraph{Overall alignment and ranking.}
Figure~\ref{fig:overall-macro} shows macro agreement with human consensus across the 30 dimensions. \texttt{claude-sonnet} leads with $0.31$, followed by \texttt{openai-o4-mini} ($0.29$), \texttt{gpt-4.1} ($0.27$), \texttt{gemini-2.5-pro} ($0.25$), \texttt{grok-2-vision} ($0.23$), \texttt{qwen2.5-vl} ($0.21$), and \texttt{llama-4-maverick} ($0.16$).
To isolate multi-label behavior, Figure~\ref{fig:overall-jaccard} reports mean Jaccard overlap with the human consensus set. The same two models occupy the top positions, with \texttt{claude-sonnet} at $0.48$ and \texttt{openai-o4-mini} at $0.45$.

\begin{figure}[H]
  \centering
  \includegraphics[width=\linewidth,height=0.65\textheight,keepaspectratio]{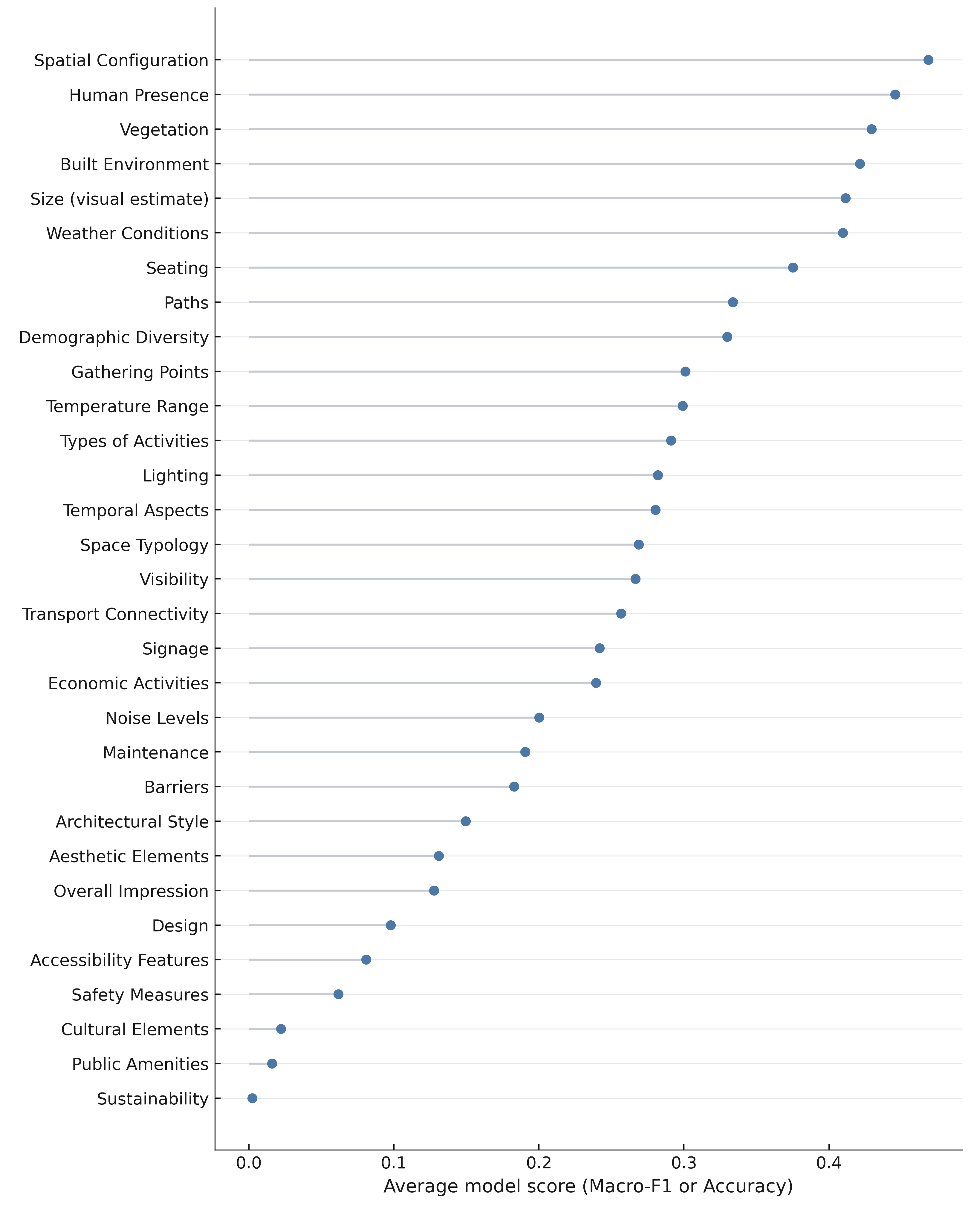}
  \caption{Difficulty by dimension. Each point is the mean model score for the dimension (accuracy or Jaccard depending on the item type).}
  \Description{Lollipop chart ranking dimensions. Easiest include Spatial Configuration, Human Presence, Vegetation, Built Environment, Size; hardest include Sustainability, Public Amenities, Cultural Elements.}
  \label{fig:difficulty-lollipop}
\end{figure}

\paragraph{Which dimensions are easier or harder?}
Figure~\ref{fig:difficulty-lollipop} ranks dimensions by average model score. The top of the list includes \emph{Spatial Configuration}, \emph{Human Presence}, \emph{Vegetation}, \emph{Built Environment}, and \emph{Size (visual estimate)} with average scores in the $0.39$--$0.47$ range. At the other extreme, \emph{Sustainability}, \emph{Public Amenities}, and \emph{Cultural Elements} are near zero, followed by \emph{Safety Measures} and \emph{Accessibility Features}. These patterns are broadly consistent with visual observability: readily visible or structural properties are recovered more reliably than diffuse or rare concepts.

\begin{figure}[H]
  \centering
  \includegraphics[width=\linewidth,height=0.65\textheight,keepaspectratio]{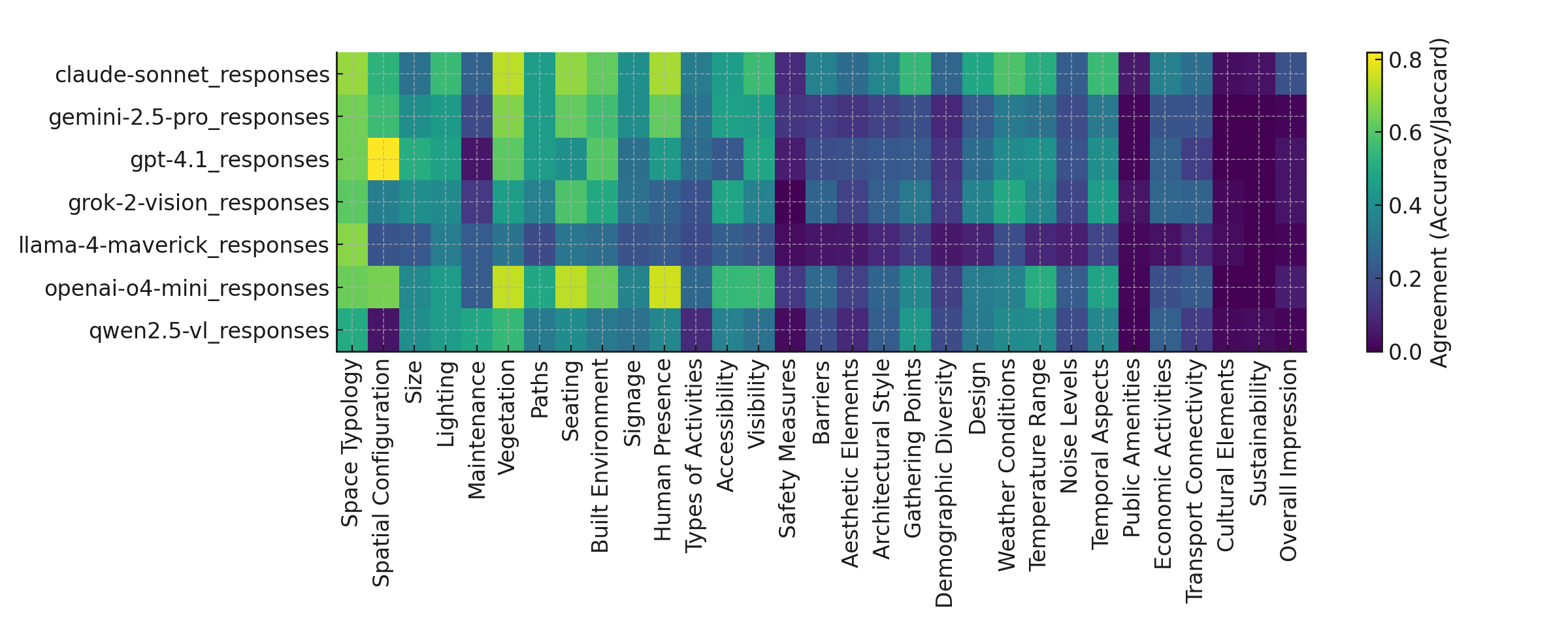}
  \caption{Agreement by dimension and model. Warmer colors indicate higher agreement with human consensus (accuracy for single-choice, Jaccard for multi-label).}
  \Description{Heatmap showing higher agreement on Space Typology, Vegetation, Seating; lower on Sustainability and Overall Impression across models.}
  \label{fig:heatmap}
\end{figure}

\paragraph{Agreement structure across models and dimensions.}
The heatmap in Figure~\ref{fig:heatmap} shows shared strengths and weaknesses across models. Most systems perform well on \emph{Space Typology}, \emph{Vegetation}, and \emph{Seating}, while struggling on \emph{Sustainability} and \emph{Overall Impression}. Model-specific differences are visible but modest relative to the cross-dimension structure, suggesting that the schema itself partitions tasks into relatively tractable and relatively difficult categories under zero-shot prompting.

\begin{figure}[t]
  \centering
  \includegraphics[width=\linewidth,height=0.65\textheight,keepaspectratio]{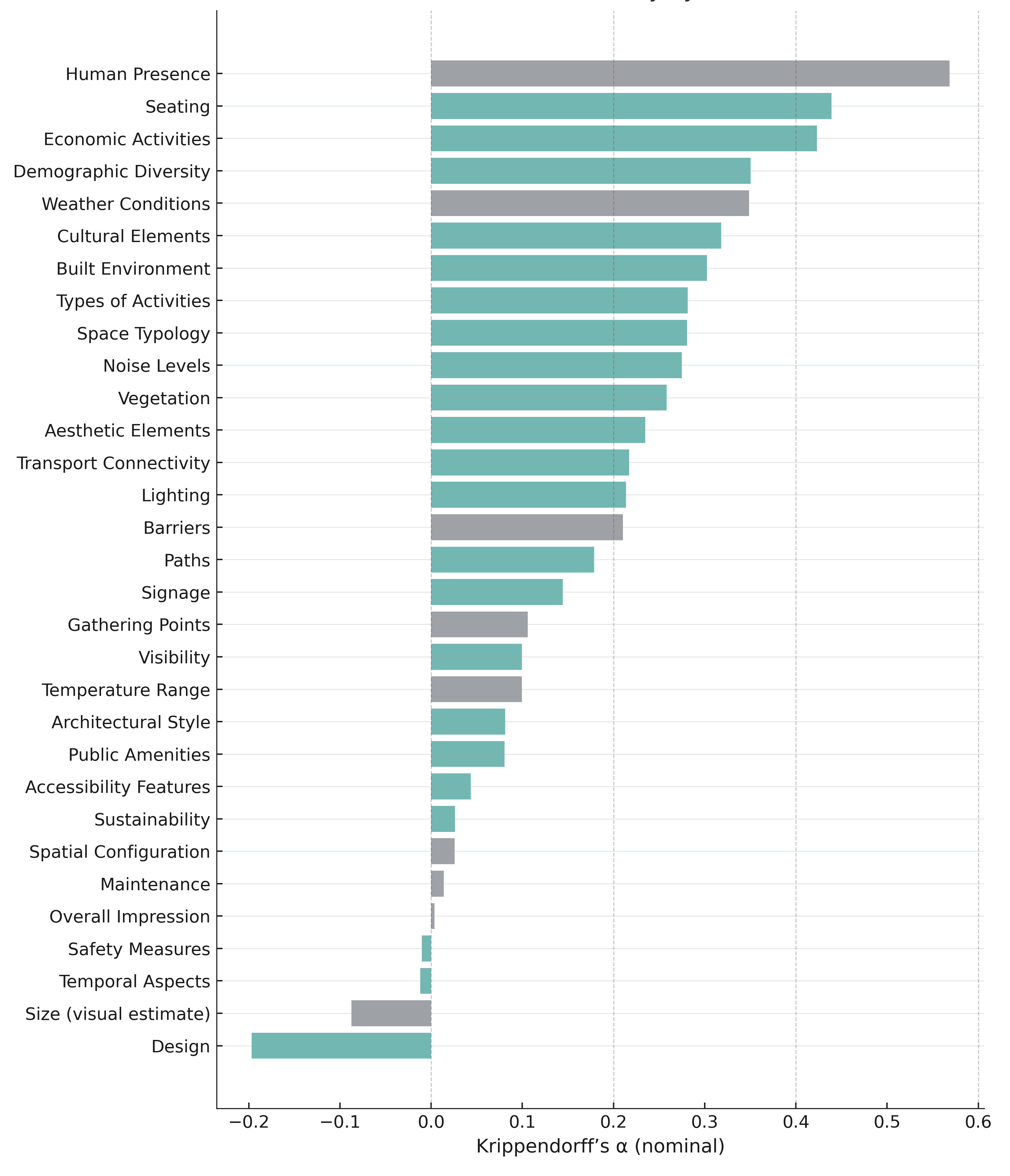}
  \caption{Inter-annotator reliability by dimension. Bars show Krippendorff's $\alpha$ (nominal).}
  \Description{Bar chart of Krippendorff's alpha per dimension. Highest for Human Presence, Seating, Economic Activities; lowest for Design, Size, Temporal Aspects.}
  \label{fig:iaa-alpha}
\end{figure}

\begin{figure}[htbp]
  \centering
  \includegraphics[width=\linewidth,height=0.40\textheight,keepaspectratio]{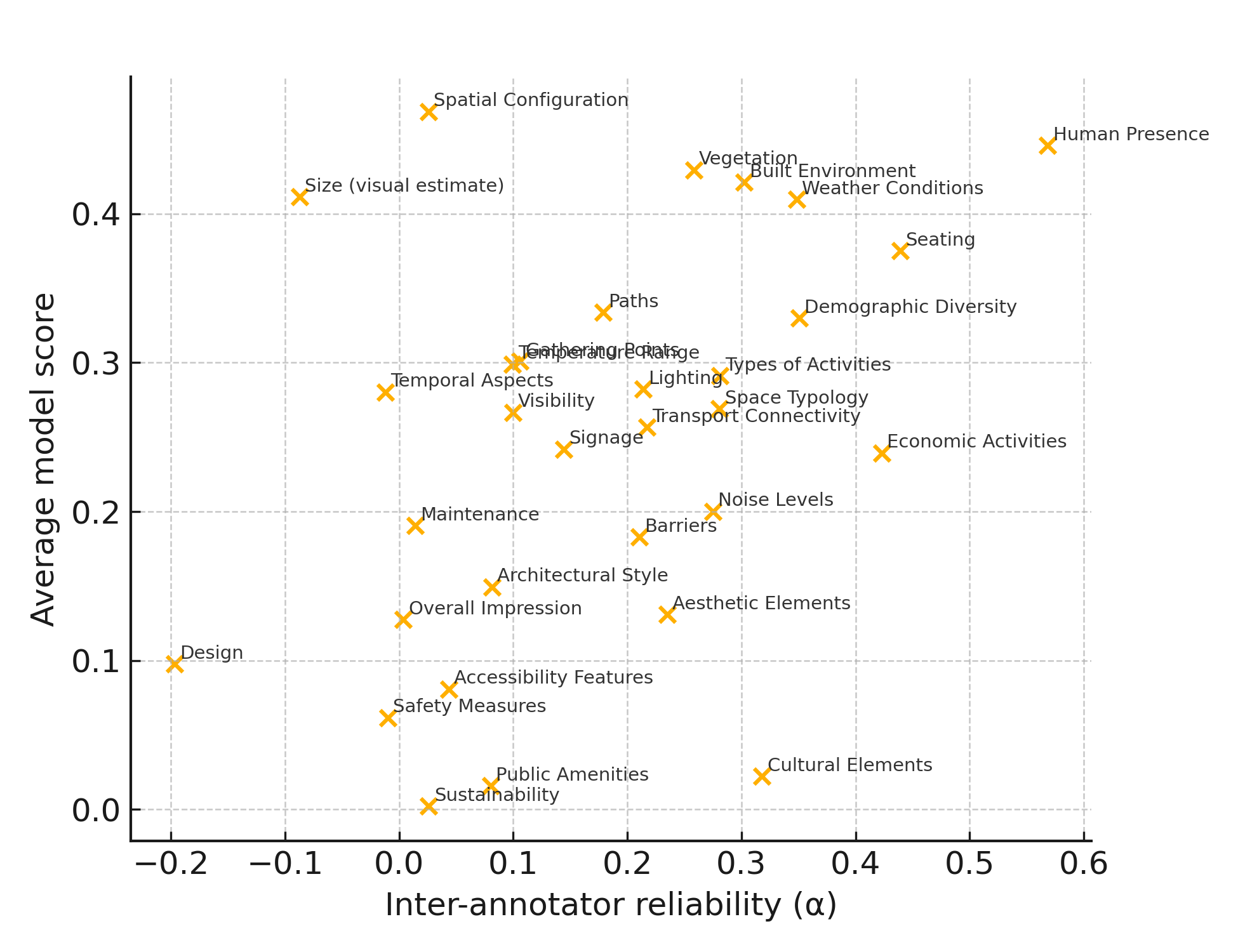}
  \caption{Relationship between human reliability (Krippendorff's $\alpha$) and average model score by dimension.}
  \Description{Scatter plot with a positive trend: higher human agreement correlates with higher model scores; outliers highlighted in text.}
  \label{fig:iaa-scatter}
\end{figure}

\paragraph{Human reliability and model performance.}
Two complementary views relate human reliability to model results. The bar chart in Figure~\ref{fig:iaa-alpha} shows Krippendorff's $\alpha$ by dimension. Human agreement is highest for \emph{Human Presence}, \emph{Seating}, and \emph{Economic Activities}; it is lowest or unstable for \emph{Design}, \emph{Size (visual estimate)}, and \emph{Temporal Aspects}. The scatter in Figure~\ref{fig:iaa-scatter} plots average model score against inter-annotator reliability. The upward trend indicates that models perform better where humans agree more. Outliers are informative: \emph{Economic Activities} has relatively high human agreement yet only middling model scores, while \emph{Spatial Configuration} achieves high model scores despite moderate human agreement, possibly because models rely on strong global layout priors even where people hesitate between open and semi-enclosed labels.

\begin{figure}[htbp]
  \centering
  \includegraphics[width=\linewidth,height=0.35\textheight,keepaspectratio]{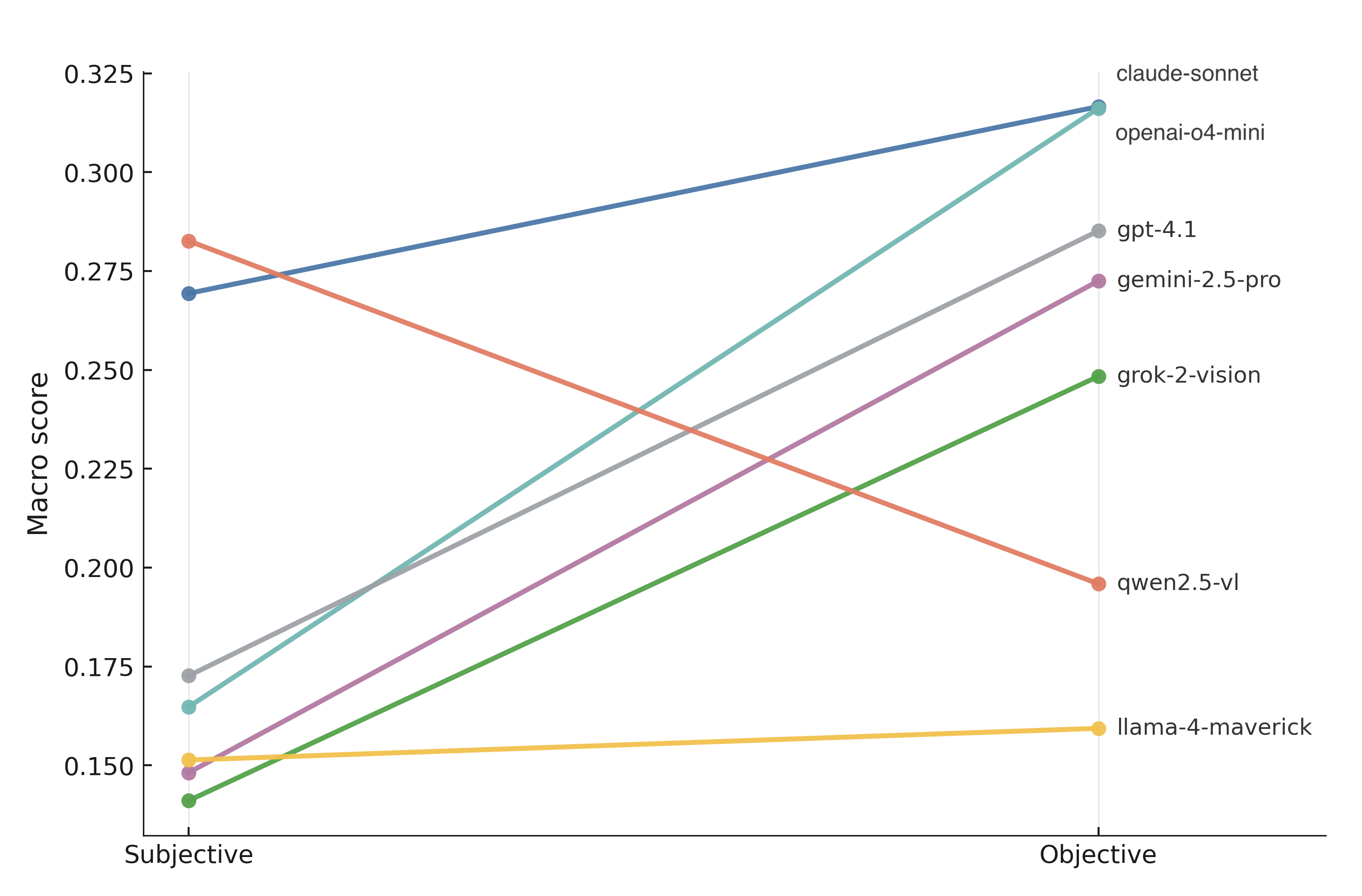}
  \caption{Subjective versus objective performance by model. Each line connects a model's macro score on the subjective subset to its macro score on the objective subset.}
  \Description{Slope chart indicating most models do better on objective dimensions; differences are largest for openai-o4-mini and gpt-4.1.}
  \label{fig:subj-obj}
\end{figure}

\paragraph{Subjective versus objective dimensions.}
We partition the schema into subjective and objective subsets and compute model scores per subset. Figure~\ref{fig:subj-obj} shows that most models improve on objective items. The difference is largest for \texttt{openai-o4-mini} and \texttt{gpt-4.1}. Two models buck the trend: \texttt{qwen2.5-vl} performs slightly worse on objective items, and \texttt{llama-4-maverick} exhibits very small differences. These divergences could reflect differences in visual backbone training or prompting sensitivity.

\begin{figure}[htbp]
  \centering
  \includegraphics[width=\linewidth,height=0.35\textheight,keepaspectratio]{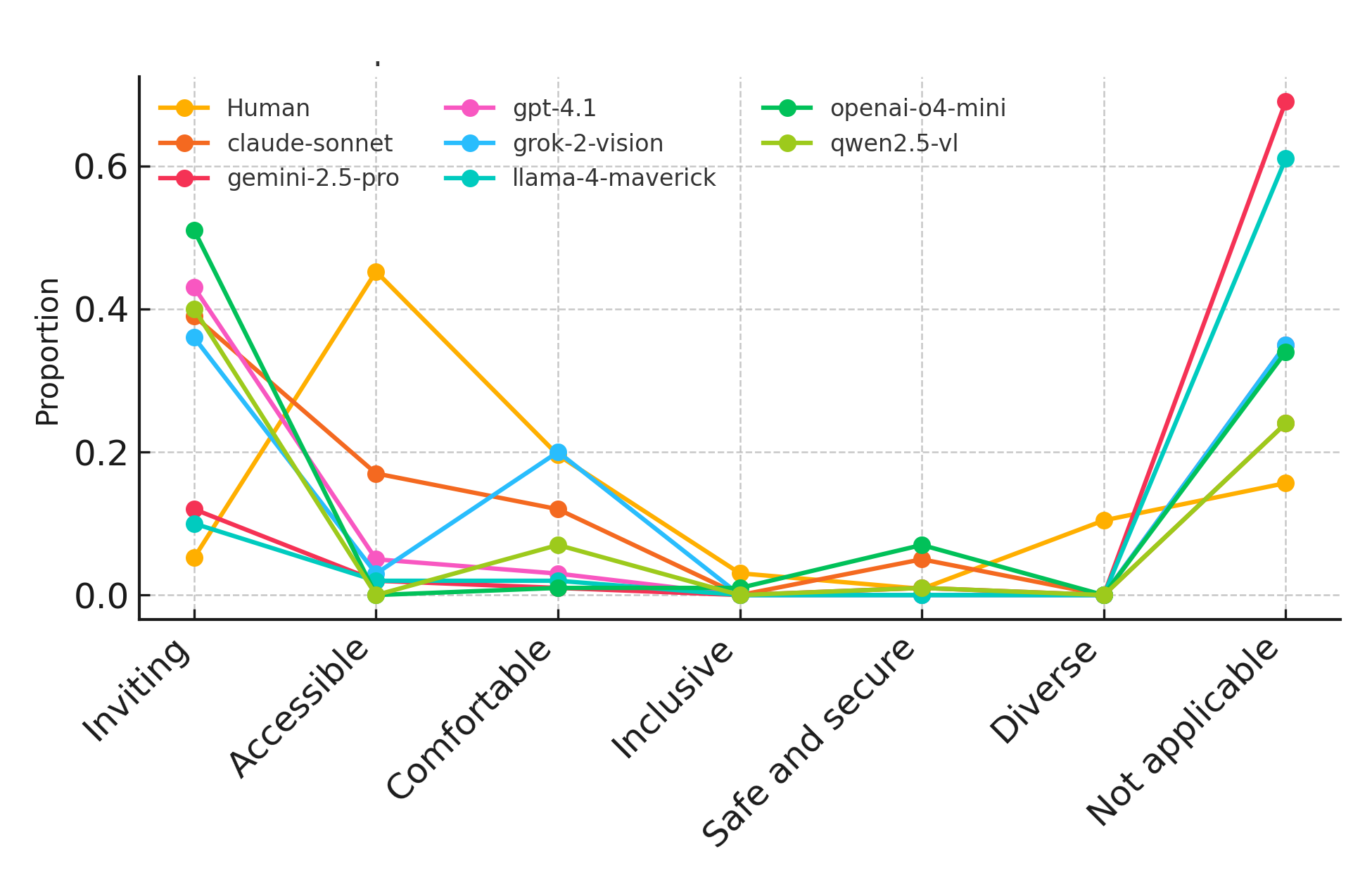}
  \caption{Overall Impression: distribution across annotators and models. Curves indicate the proportion of images assigned to each category.}
  \Description{Distribution plot comparing humans and models; several models over-select “Not applicable” and under-select “Accessible”.}
  \label{fig:overall-impression}
\end{figure}

\paragraph{Distributional mismatch on subjective appraisals.}
Beyond accuracy, distributional differences illuminate qualitative mismatches. Figure~\ref{fig:overall-impression} plots the proportion of labels selected for \emph{Overall Impression}. Human annotators most often selected \emph{Accessible} and \emph{Comfortable}, whereas several models over-produce \emph{Not applicable} and under-produce \emph{Accessible}. This mismatch indicates that models do not simply make random errors; rather, they follow different priors over subjective categories.

\paragraph{Image source effects.}
Across models we observe modest but consistent decreases on synthetic images relative to photographs. The gap is small compared to cross-dimension variation and does not change the ordering of models. Qualitative inspection suggests that synthetic scenes sometimes contain idealized plazas and abundant seating, which humans and models both interpret differently than in photographs of typical residential streets. We provide per-source breakdowns in the supplementary tables.

\section{Discussion}

The benchmark suggests a clear stratification by observability. Dimensions grounded in readily visible structure, such as spatial configuration or presence of vegetation, are detected more consistently even in a zero-shot setting. By contrast, low-scoring dimensions combine two challenges: the required evidence is either rare in our panels (\emph{Sustainability}, \emph{Public Amenities}) or it involves diffuse cultural and affective interpretation (\emph{Cultural Elements}, \emph{Overall Impression}). These results echo observations in urban studies that subjective impressions are plural and context dependent, and they support presenting distributions and uncertainty rather than single labels \citep{Mushkani2025ICML_LIVS}.

The relationship between human reliability and model performance suggests that VLMs are sensitive to the intrinsic clarity of a dimension. Where people agree strongly, models typically do too; where people disagree, models also struggle. This argues for evaluation protocols that place model scores alongside human reliability rather than against an assumed ground truth. It also creates an opportunity: identifying dimensions with high human agreement but modest model scores, such as \emph{Economic Activities}, can guide targeted data collection or prompt engineering.

The distributional analysis of \emph{Overall Impression} demonstrates that differences are not only about accuracy. Several models fall back on \emph{Not applicable} more often than humans do, a conservative strategy that may be reasonable when the visual evidence is weak but that limits usefulness in participatory settings. Interfaces that allow models to express calibrated uncertainty could mitigate this issue.

Finally, the modest real-versus-synthetic gap is encouraging for data augmentation. Synthetic panels helped expose rare combinations (for example, sunlit plazas with diverse activities). They also risk imprinting stylistic biases from the generator. Future work could probe stronger domain shifts and include a more explicit assessment of synthetic artifacts.

\section{Ethics and Privacy}
All participants provided informed consent and were compensated. Annotations are released only in aggregate form with no personal data. Photographs were captured by the authors in public spaces; faces and license plates were automatically blurred and manually verified prior to use. Synthetic images were generated with Stable Diffusion XL.

\section{Limitations}

The dataset is intentionally small to enable careful documentation, community partnership, and reliability analysis. This limits statistical power for subgroup analyses and restricts the diversity of Montreal contexts. Annotations reflect a specific group of community members; other groups could disagree. We use a deterministic parser for French--English normalization and CSV extraction; although validated, any remaining mapping errors would underrate models. Our evaluation is strictly zero-shot. Light-weight tuning with local data or multi-turn interaction may yield higher agreement but would assess a different capability. Finally, while we avoid personally identifying information in images and release only de-identified annotations, any work that links urban perception with place requires caution to avoid stigmatizing neighborhoods or groups.

\section{Implications for practice and research}

For practice, present-day VLMs may assist with factual components of streetscape audits, such as identifying seating, vegetation, and typology, and can pre-annotate images for human review. They are less dependable for subjective appraisals. Tools should therefore surface both model predictions and human reliability for each dimension, display uncertainty, and allow community members to contest labels.

For research, the benchmark suggests two directions. The first is methodological: evaluation should normalize for inter-annotator variability, report both accuracy and distributional fit, and separate subjective from objective items. The second is technical: future models could incorporate structured visual geometry for spatial tasks and learn to express calibrated uncertainty for subjective appraisals. Participatory co-design with local organizations can help align model outputs with situated values.

\section{Conclusion}

We introduced a community-grounded benchmark for evaluating vision--language models on urban perception using Montreal street-level scenes. The analysis shows stronger alignment on objective dimensions, weaker alignment on subjective appraisals, and a positive relationship between human reliability and model performance; we also observe a modest but consistent performance drop on photorealistic synthetic scenes. Taken together, these results provide baseline yardsticks for seven contemporary VLMs under a strict zero-shot protocol and a documented pipeline that emphasizes determinism and transparency.

Three methodological takeaways emerge. \emph{First, observability matters:} dimensions with readily visible, structural evidence (e.g., vegetation, spatial configuration) are recovered far more reliably than diffuse or rare concepts (e.g., sustainability, cultural elements). \emph{Second, reliability matters:} model scores co-vary with inter-annotator agreement, arguing for evaluation that reports model alignment \emph{alongside} human reliability rather than against a presumed single ground truth. \emph{Third, distribution matters:} on subjective appraisals, models exhibit distinct priors (e.g., overusing \texttt{Not applicable}); surfacing calibrated uncertainty and presenting distributions rather than single labels can make these systems more useful in participatory contexts.

For practice, the benchmark suggests that present-day VLMs can help pre-annotate factual components of streetscape audits (seating, typology, vegetation) and support triage for human review, but they should not be relied upon as arbiters of subjective qualities. Interfaces should therefore (i) disclose per-dimension human reliability, (ii) expose model abstention and confidence, and (iii) allow community members to contest or revise labels. These design choices align with participatory goals and reduce the risk of reifying model priors as facts.

For research, several extensions follow naturally. Scaling beyond 100 scenes to additional neighborhoods, seasons, and cities---and engaging a broader set of community groups and languages---would test generalization and illuminate place-specific priors. Richer targets (e.g., pairwise preferences, ordinal ratings, or small, structured rationales) could probe whether models learn \emph{why} a label applies, not just \emph{which} one. Technique-wise, future systems may benefit from (i) uncertainty-aware objectives that calibrate abstention, (ii) better spatial reasoning via explicit geometric cues, (iii) multi-image or multi-turn protocols that let models reconcile ambiguous evidence, and (iv) analyses of domain shift beyond photorealism (e.g., night scenes, weather, seasonality). Because perception is situated, future evaluations should also include value-sensitive comparisons and fairness checks across communities.

We release prompts, normalization mappings, and a deterministic harness to support reproducible comparisons and community extension. Our aim is a living benchmark that helps the urban AI community track progress on human–AI alignment, encourages uncertainty-aware reporting, and keeps participatory values at the center of street-scale analysis.

\begin{table}[H]
\centering
\caption{Summary of overall model alignment. Left: macro score across all 30 dimensions (accuracy for single-choice; Jaccard for multi-label). Right: mean Jaccard restricted to the multi-label subset. Values match Figures~\ref{fig:overall-macro} and~\ref{fig:overall-jaccard}.}
\label{tab:summary-macro}
\begin{tabular}{lcc}
\toprule
\multirow{2}{*}{Model} & Macro (all dims) & Mean Jaccard \\
 & (Acc/Jaccard) & (multi-label only) \\
\midrule
claude\texttt{-}sonnet & 0.31 & 0.48 \\
openai\texttt{-}o4\texttt{-}mini & 0.29 & 0.45 \\
gpt\texttt{-}4.1 & 0.27 & 0.35 \\
gemini\texttt{-}2.5\texttt{-}pro & 0.25 & 0.36 \\
grok\texttt{-}2\texttt{-}vision & 0.23 & 0.39 \\
qwen2.5\texttt{-}vl & 0.21 & 0.28 \\
llama\texttt{-}4\texttt{-}maverick & 0.16 & 0.27 \\
\bottomrule
\end{tabular}
\end{table}

\bibliographystyle{plainnat}
\bibliography{references}

\appendix

\section{Perception grid}
\label{app:grid}

The benchmark uses \textbf{30} dimensions.
\vspace{1em}

\begin{small}
\setlength{\LTpre}{0pt}\setlength{\LTpost}{0pt}
\begin{longtable}{p{0.22\linewidth} p{0.68\linewidth} p{0.08\linewidth}}
\caption{Perception grid: dimensions, allowed values, and type.}
\label{tab:perception-grid}\\
\toprule
\textbf{Dimension} & \textbf{Allowed values} & \textbf{Type}\\
\midrule
\endfirsthead
\toprule
\textbf{Dimension} & \textbf{Allowed values} & \textbf{Type}\\
\midrule
\endhead
\midrule
\multicolumn{3}{r}{\emph{Continued on next page}}\\
\midrule
\endfoot
\bottomrule
\endlastfoot

Space Typology & \texttt{Park; Street; Square; Courtyard; Garden; Waterfront; Public plaza; Alley; Playground; Not applicable} & multiple\\

Spatial Configuration & \texttt{Open; Enclosed; Semi-enclosed; Structured; Organic; Not applicable} & single\\

Size (visual estimate) & \texttt{Small (<500 m²); Medium (500–2000 m²); Large (>2000 m²); Not applicable} & single\\

Lighting & \texttt{Natural lighting; Artificial lighting; Well lit; Poorly lit; Shaded areas; Not applicable} & multiple\\

Maintenance & \texttt{Clean; Dirty; Well maintained; Neglected; Recently renovated; Not applicable} & single\\

Vegetation & \texttt{Trees present; Too much greenery; Little greenery; Grass present; Bushes present; Flower beds present; No vegetation; Not applicable} & multiple\\

Paths & \texttt{Paved paths present; Unpaved paths present; Wide paths present; Narrow paths present; Linear paths present; Curved paths present; Intersecting paths present; Dead-end paths present; Not applicable} & multiple\\

Seating & \texttt{Benches present; Chairs present; Picnic tables present; Custom seats present; Movable seats present; No seating; Not applicable} & multiple\\

Built Environment & \texttt{Modern buildings present; Historic buildings present; Residential buildings present; Commercial buildings present; Mixed-use buildings present; Vacant lots present; Not applicable} & multiple\\

Signage & \texttt{Informational signs present; Decorative signs present; Directional signs present; Interactive signs present; No signage; Not applicable} & multiple\\

Human Presence & \texttt{Crowded (>50 people); Moderately populated (20–50 people); Sparsely populated (<20 people); Empty; Not applicable} & single\\

Types of Activities & \texttt{Recreational activities present; Leisure activities present; Commercial activities present; Transportation activities present; Cultural activities present; Social activities present; Sports activities present; Religious activities present; Not applicable} & multiple\\

Accessibility Features & \texttt{Ramps present; Handrails present; Tactile paving present; Elevators present; Wide entrances present; Accessible restrooms present; No accessibility features; Not applicable} & multiple\\

Visibility & \texttt{Clear sight lines; Obstructed views present; Panoramic views present; Hidden corners present; Not applicable} & multiple\\

Safety Measures & \texttt{Surveillance cameras present; Security personnel present; Safety lighting; Emergency exits present; Safety signs present; Fences present; Walls present; Not applicable} & multiple\\

Barriers & \texttt{Physical barriers present (fences, walls); Natural barriers present (rivers, hills); No barriers; Not applicable} & single\\

Aesthetic Elements & \texttt{Bright colours present; Dark colours present; Monochrome elements present; Murals present; Sculptures present; Street art present; Water features present; No decorative elements; Not applicable} & multiple\\

Architectural Style & \texttt{Traditional buildings present; Contemporary buildings present; Eclectic buildings present; Vernacular buildings present; Post-modern buildings present; Brutalist buildings present; Not applicable} & multiple\\

Gathering Points & \texttt{Central gathering point present; Edge gathering points present; Gathering points near monuments present; Informal gathering points present; No gathering points; Not applicable} & single\\

Observed Group Diversity & \texttt{Variety in group sizes; Presence of family groups; Presence of mixed-age groups; Presence of mobility aids; Not applicable} & multiple\\

Inclusive Design Features & \texttt{Wheelchair-accessible features present; Braille signage present; Multilingual signs present; Gender-neutral restrooms present; Adapted play equipment present; No design features; Not applicable} & multiple\\

Weather Conditions & \texttt{Sunny; Rainy; Snowy; Cloudy; Windy; Foggy; Not applicable} & single\\

Temperature Range & \texttt{Hot (>30 °C); Warm (20–30 °C); Cool (10–20 °C); Cold (<10 °C); Not applicable} & single\\

Noise Levels & \texttt{Quiet; Moderate; Loud; Traffic noise present; Construction noise present; Natural sounds present; Not applicable} & multiple\\

Temporal Aspects & \texttt{Daytime; Night; Weekday; Weekend; Seasonal variations; Not applicable} & single\\

Public Amenities & \texttt{Restrooms present; Water fountains present; Information kiosks present; Trash bins present; Play areas present; Fitness equipment present; Not applicable} & multiple\\

Economic Activities & \texttt{Street vendors present; Markets present; Shops present; Cafés present; No commercial activities; Not applicable} & multiple\\

Transport Connectivity & \texttt{Public transport access present; Bicycle lanes present; Pedestrian paths present; Parking spaces present; Carpool points present; Not applicable} & multiple\\

Cultural Elements & \texttt{Historic monuments present; Monuments present; Culturally significant features present; Public art installations present; Not applicable} & multiple\\

Sustainability & \texttt{Recycling bins present; Green building features present; Use of renewable energy present (e.g., solar panels); Water conservation measures present; Not applicable} & multiple\\

Overall Impression & \texttt{Inviting; Accessible; Comfortable; Inclusive; Safe and secure; Diverse; Cannot judge; Not applicable} & single\\

\end{longtable}
\end{small}

\paragraph{Tokenization note.}
The harness expects a \emph{single CSV line} per image with \emph{30} comma-separated fields, one per dimension above; multi-label selections are joined with semicolons and no spaces (e.g., \texttt{Natural lighting;Well lit}). The output CSV written to disk adds two columns around this line-level response: a leading \texttt{Image\_ID} and a trailing \texttt{Comments} field used for parser diagnostics. The model never outputs \texttt{Image\_ID}.

\section{Prompt and parsing details}
\label{app:prompt}

The inference script provides a deterministic system prompt that: (1) lists the 30 columns in the exact order shown in Table~\ref{tab:perception-grid}; (2) requires \emph{only} a single CSV line with 30 fields and no header or commentary; (3) instructs models to use \texttt{Not applicable} when unclear; and (4) specifies the semicolon rule for multi‑label items. Parsing is rule‑based: replies are split on commas (with a CSV fallback), trimmed and mapped to the codebook tokens, and padded/truncated to the expected length. Any non‑conforming reply is captured in \texttt{Comments} and excluded from scoring.

\section{Implementation notes for reproducibility}
\label{app:impl}

To ensure the models truly see pixels and that results are repeatable, the benchmark harness:

\begin{enumerate}[leftmargin=*, itemsep=0.25em]
  \item embeds local images as \emph{base‑64 data URIs} in the chat message (no remote fetches);
  \item sets \texttt{temperature=0}, \texttt{top\_p=1}, fixes max tokens, and logs model version strings and run timestamps;
  \item uses exponential retry/backoff for transient API errors and logs the first 120 characters of each raw reply at \texttt{DEBUG};
  \item discovers images under \texttt{p1}–\texttt{p10} panel folders (expecting 100 files);
  \item enforces the one‑line CSV contract and a deterministic post‑processor that writes \texttt{Image\_ID + 30 fields + Comments} to disk.
\end{enumerate}

\section{Image source effects (real vs.\ synthetic)}
\label{app:real-vs-synth}

\begin{figure}[H]
  \centering
  \includegraphics[width=\linewidth,height=0.35\textheight,keepaspectratio]{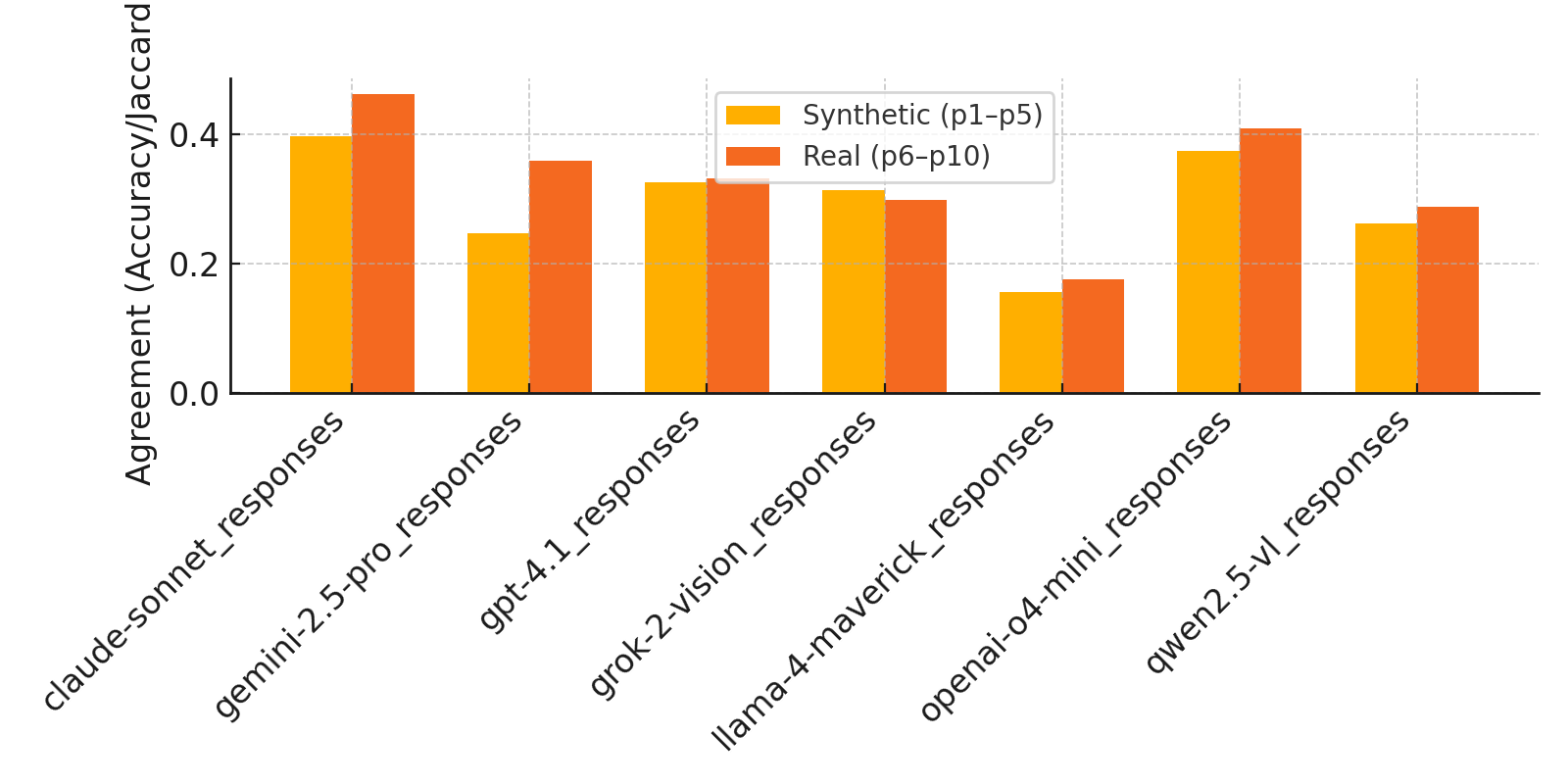}
  \caption{Model agreement by image source. Orange: real photos ($p6$–$p10$). Gold: synthetic renders ($p1$–$p5$).}
  \Description{Bar chart comparing agreement of seven VLMs on real photographs versus synthetic images; real is modestly higher across models but rankings are unchanged.}
  \label{fig:real-vs-synth}
\end{figure}

Panels $p1$–$p5$ consist of photorealistic \emph{synthetic} scenes and $p6$–$p10$ are \emph{real} photographs (50 images each). Figure~\ref{fig:real-vs-synth} breaks down agreement by source. We observe modest, consistent decreases on synthetic images across models, but the gap is small compared with cross‑dimension variation and does not change the ranking.

\section{Prompt template}
\label{app:prompt-template}

\begin{tcolorbox}[colback=gray!10, colframe=gray, title=Prompt contract (model-facing)]
\begin{lstlisting}[style=mystyle, language=Python]
def build_system_prompt() -> str:
    lines = [
        "You are an expert urban-perception assessor.",
        "Return ONLY a single CSV line (no header) with 30 comma-separated fields,",
        "one field per dimension, in the exact order listed below.",
        "For multi-select dimensions, join choices with semicolons (no spaces).",
        "If evidence is unclear or absent, output Not applicable.",
        "Do NOT include an image id, quotes, or commentary—just the CSV line.",
        "",
        "Example (for the Lighting dimension only, which is column 4):",
        ",,,Natural lighting;Well lit,,,,,,,,,,,,,,,,,,,,,,,,,,,,,",
        "",
        "Column order and allowed values:",
    ]
    for i, dim in enumerate(GRID, 1):
        flag = " (multiple)" if dim.multiple else " (single)"
        lines.append(f"{i}. {dim.name}{flag}: " + "; ".join(dim.variables))
    lines.append("Return just the CSV line—nothing else.")
    return "\n".join(lines)
\end{lstlisting}
\end{tcolorbox}

\end{document}